\pdfoutput=1
\documentclass{article}
\usepackage[letterpaper,top=2cm,bottom=2cm,left=3cm,right=3cm,marginparwidth=1.75cm]{geometry}

\usepackage[utf8]{inputenc}
\usepackage[T1]{fontenc}
\usepackage{hyperref}[]
\usepackage{url}
\usepackage{booktabs}
\usepackage{natbib}
\usepackage{amsfonts}
\usepackage{nicefrac}
\usepackage{microtype}
\usepackage{xspace}
\usepackage{xcolor}
\usepackage{amsmath}
\usepackage{amssymb}
\usepackage[section]{placeins}
\usepackage{graphicx}
\usepackage{natbib}
\usepackage{booktabs}
\usepackage{xcolor}
\usepackage{colortbl} %
\usepackage{soul} %
\usepackage{tabularx}
\usepackage{stfloats}
\usepackage{algorithm,algorithmicx}
\usepackage[noend]{algpseudocode}
\usepackage{algcompatible}
\graphicspath{{./figures/}}
\usepackage{multirow}
\usepackage[most]{tcolorbox}
\usepackage{mathtools}
\usepackage{shortvrb}
\usepackage{subcaption}
\usepackage{pifont}%

\usepackage{amsmath,amsfonts,bm}

\def\eqref#1{equation~\ref{#1}}

\def\1{\bm{1}}

\def\rvx{{\mathbf{x}}}

\def\vtheta{{\bm{\theta}}}

\DeclareMathAlphabet{\mathsfit}{\encodingdefault}{\sfdefault}{m}{sl}
\SetMathAlphabet{\mathsfit}{bold}{\encodingdefault}{\sfdefault}{bx}{n}

\newcommand{\E}{\mathbb{E}}

\newcommand{\cmark}{\ding{51}}%
\newcommand{\xmark}{\ding{55}}%

\definecolor{lightgrey}{HTML}{dcdbdb}
\definecolor{lightblue}{HTML}{E8F0FE}
\definecolor{gray}{HTML}{9aa0a6}
\definecolor{lightpink}{HTML}{F48FB1}
\definecolor{lightred}{HTML}{FFCBC9}
\definecolor{lightcyan}{HTML}{80DEEA}

\newtcolorbox{mybox}[2][]
  {colback = black!5!white, colframe = black!75!black, fonttitle = \bfseries,
    colbacktitle = black!100!black, enhanced, 
    attach boxed title to top left={yshift=-2.2mm,xshift=4mm},
    title=#2,#1}

\newcommand{\base}{Dream-Coder-7B\xspace}
\newcommand{\instruct}{Dream-Coder-7B-Instruct\xspace}

\usepackage{soul}

\setcitestyle{round}

\newcommand{\github}{\raisebox{-1.5pt}{\includegraphics[height=1.05em]{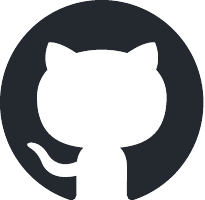}}\xspace}
\newcommand{\huggingface}{\raisebox{-1.5pt}{\includegraphics[height=1.05em]{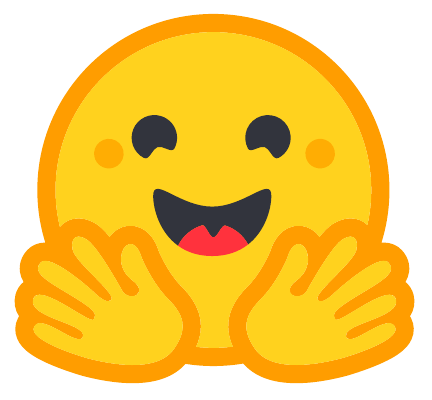}}\xspace}

\title{Dream-Coder 7B: An Open Diffusion Language Model for Code}

\author{%
  \textbf{Zhihui Xie}$^1$\thanks{Equal contribution} \quad
  \textbf{Jiacheng Ye}$^1$\footnotemark[1] \quad
  \textbf{Lin Zheng}$^1$\footnotemark[1] \quad
  \textbf{Jiahui Gao}$^2$\footnotemark[1] \and
  \textbf{Jingwei Dong}$^1$ \quad
  \textbf{Zirui Wu}$^1$ \quad
  \textbf{Xueliang Zhao}$^1$ \quad
  \textbf{Shansan Gong}$^1$ \and
  \textbf{Xin Jiang}$^2$ \quad
  \textbf{Zhenguo Li}$^2$\thanks{Core advising} \quad
  \textbf{Lingpeng Kong}$^1$\footnotemark[2] \\
  \vspace{0.5em}
  $^1$The University of Hong Kong \quad $^2$Huawei Noah's Ark Lab \\
  \github\href{https://github.com/DreamLM/Dream-Coder}{\textbf{DreamLM/Dream-Coder}} \quad \huggingface\href{https://huggingface.co/collections/Dream-org/dream-coder-7b-68761cfa0e218f0776a84ee7}{\textbf{Dream-Coder}}
  \quad \huggingface\href{https://huggingface.co/spaces/ZiruiWu/Dream-Coder-Instruct-7B}{\textbf{Demo}}
}

\date{July 15, 2025}

\begin{document}

\maketitle
\begin{abstract}
We present Dream-Coder 7B, an open-source discrete diffusion language model for code generation that exhibits emergent any-order generation capabilities. Unlike traditional autoregressive (AR) models that decode strictly left-to-right, Dream-Coder 7B adaptively determines its decoding strategy based on the coding task: sketch-first generation for complex algorithms, left-to-right generation for straightforward completions, and interleaved reasoning generation for code understanding tasks. We adapt a pretrained AR checkpoint to a discrete diffusion frameworks with a continuous-time weighted cross-entropy objective. Our post-training recipe comprises (i) supervised fine-tuning, where we mitigate padding pathologies via random truncation and a padding penalty to improve sample efficiency and stabilize generation; and (ii) reinforcement learning with verifiable rewards over a curated high-quality prompt set drawn from open-source datasets, using a tailored reinforcement learning recipe for diffusion language models. The resulting Dream-Coder 7B Instruct attains 21.4\% pass@1 on LiveCodeBench (2410--2505) and demonstrates competitive performance on HumanEval, MBPP, BigCodeBench, and CRUXEval. We release \base and \instruct checkpoints, training recipes, preprocessing pipelines, and inference code to facilitate reproducibility and further research.
\end{abstract}

\section{Introduction}
\label{sec:intro}

The rapid advancement of large language models has fundamentally transformed software development, with AI-powered coding assistants becoming indispensable tools for millions of developers worldwide~\citep{claude_4,openAI_o3_o4_mini,comanici2025gemini}. The current landscape is dominated by autoregressive (AR) models that generate code through sequential left-to-right token prediction. While these models have demonstrated remarkable capabilities in code completion and generation, a central question remains: What architectural paradigms might define the next generation of code generation models?

This question becomes increasingly critical as AR models exhibit inherent limitations in complex programming scenarios~\citep{bachmann2024pitfalls}. Consider code refactoring, where developers need to simultaneously modify function signatures, update call sites, and maintain consistency across entire codebases. Or debugging scenarios where understanding requires bidirectional context flow and iterative hypothesis testing. These tasks demand global program understanding, flexible infilling, and multi-step refinement—capabilities where sequential left-to-right generation struggles. The rigid token-by-token AR paradigm constrains holistic reasoning about program structure and semantics.

Discrete diffusion models offer a compelling alternative for code generation by beginning with completely corrupted sequences and gradually denoising them through multiple refinement steps~\citep{austin2021structured,zheng2023reparameterized}. This architectural shift unlocks several theoretical advantages that directly address autoregressive limitations. The bidirectional nature of diffusion enables richer contextual modeling by incorporating information from all positions simultaneously, leading to more coherent and globally consistent code generation~\citep{li2022diffusion}. Furthermore, the iterative refinement process naturally supports controllable generation and provides flexible quality-speed trade-offs through adjustable inference steps, opening new dimensions for test-time optimization~\citep{snell2024scaling} that complement existing techniques.

Recent developments have begun to demonstrate the practical viability of diffusion models for code generation. While general-purpose diffusion language models like LLaDA~\citep{nie2025large} and Dream 7B~\citep{dream2025} have established the foundation for scaled diffusion modeling, code-specific applications present unique challenges and opportunities. Commercial implementations such as Mercury Coder~\citep{khanna2025mercury} have shown promising results, while research efforts like DiffuCoder~\citep{gong2025diffucoder} have explored masked diffusion strategies specifically for programming languages. However, significant gaps remain in achieving performance parity with state-of-the-art autoregressive code models while fully leveraging the unique capabilities of diffusion-based generation.

In this work, we present \textbf{Dream-Coder 7B}, a specialized diffusion language model that bridges this performance gap while demonstrating novel capabilities inherent to the diffusion paradigm for code generation. Building upon the foundational advances of Dream 7B in general diffusion language modeling, we introduce a comprehensive training framework specifically tailored for code generation tasks. Our approach combines AR-based initialization from Qwen2.5-Coder with context-adaptive noise scheduling and novel post-training techniques including verifiable reward-driven reinforcement learning. Through extensive evaluation, we establish that diffusion models can match autoregressive performance on standard coding benchmarks while providing distinct advantages in complex programming scenarios, offering unprecedented generation flexibility through adaptive pattern selection and iterative refinement capabilities.

\begin{mybox}[colback=gray!10]{Takeaways}
    \begin{itemize}
      \item We present Dream-Coder 7B, the first open-source diffusion language model for code generation to achieve competitive performance with AR baselines while offering unique generation capabilities.
      \item We demonstrate three emergent adaptive generation patterns—sketch-first scaffolding, left-to-right completion, and interleaved reasoning—that naturally arise from diffusion training and adapt to different coding task complexities.
      \item We release the model, training recipes, preprocessing pipelines, and inference code to facilitate reproducibility and further research.
    \end{itemize}
\end{mybox}

\section{\base}
\label{sec:pretrain}

\subsection{Background: Discrete Diffusion Modeling}
Discrete diffusion models~\citep{austin2021structured,hoogeboom2021argmax,campbell2022continuous} are a class of latent variable models characterized by a forward noising process and a learned reverse denoising process. Let $\rvx_t$ denote the noised sequence at timestep $t$, with \texttt{[MASK]} tokens used to represent noise. The forward process $q(\rvx_{1:T}|\rvx_0) = \prod_{t=1}^T q(\rvx_t|\rvx_{t-1})$ progressively corrupts the original data $\rvx_0 \coloneqq \rvx$ into increasingly noisy sequences $\rvx_{1:T} \coloneqq \rvx_1, \dots, \rvx_T$. At each timestep, a noise schedule $\alpha_t$ controls the probability that each token remains unmasked. The backward process then learns to gradually denoise the masked sequence back to the original data distribution by iteratively predicting masked tokens as $t$ decreases from $T$ to $0$:
\begin{equation*}
    p_{\vtheta}(\rvx) = \sum_{\rvx_{1:T} \sim q} p(\rvx_T) \prod_{t=1}^T p_{\vtheta}(\rvx_{t-1} \mid \rvx_t),
\end{equation*}
where the denoising steps leverage full-sequence context rather than the left-context prediction of autoregressive models.

While traditional discrete diffusion employs fixed discrete timesteps $t \in \{0, \dots, T\}$, this can introduce bias by restricting $\rvx_t$ to predetermined noise levels. Following~\citet{kingma2021variational,campbell2022continuous,sahoo2024simple,shi2025simplifiedgeneralizedmaskeddiffusion,ou2024your}, we adopt a continuous-time parameterization where $t \in [0, 1]$, enabling $q(\rvx_t|\rvx_s)$ for any $0 \leq s < t \leq 1$. This allows for flexible noise schedules and sampling across arbitrary noise levels. We use absorbing-state diffusion and train with a weighted cross-entropy loss:
\begin{equation}\label{eq:train_obj}
    L(\theta) = -\E_{\rvx_0 \sim q(\rvx), t \sim \mathcal{U}(0,1), \rvx_t \sim q(\rvx_t \mid \rvx_0)} w(t) \sum_{n=1}^N \1_{[\rvx_t^n=\texttt{MASK}]} \log p_{\theta}(\rvx_0^n \mid \rvx_t),
\end{equation}
where the indicator $\1_{[\rvx_t^n=\texttt{MASK}]}$ ensures the loss is computed only on masked positions, and $w(t) \in (0,1]$ is a time-dependent reweighting term determined by the noise schedule $\alpha_t$~\citep{zheng2023reparameterized,shi2025simplifiedgeneralizedmaskeddiffusion,gong2025scalingdiffusionlanguagemodels}. This formulation provides a tractable variational upper bound on the negative log-likelihood, making it a practical and effective training objective for large-scale diffusion language models.

\subsection{Architecture}
Dream-Coder 7B is a discrete diffusion language model based on the Qwen2.5-Coder 7B model~\citep{hui2024qwen2}. 
Following \cite{gong2025scalingdiffusionlanguagemodels,dream2025}, we adopt the \emph{shift operation} strategy to adapt the autoregressive model architecture while retaining the learned knowledge from initialization. 

Specifically, unlike standard diffusion models that predict tokens at their masked positions, the shift operation allows the model to leverage its pretrained capabilities while gaining the benefits of bidirectional context and parallel generation inherent in diffusion modeling.
For details, please refer to \cite{gong2025scalingdiffusionlanguagemodels}.

\section{Pre-training}

\subsection{Objective}
The Dream-Coder 7B model is trained to predict clean tokens from their noised counterparts under a continuous-time discrete diffusion framework (Equation~\ref{eq:train_obj}). At each training step, we sample a clean sequence $\rvx_0$ from the pretraining corpus, draw a noise level $t \sim \mathcal{U}(0,1)$, and apply the forward noising process $q(\rvx_t \mid \rvx_0)$ to replace a subset of tokens with \texttt{[MASK]} according to the noise schedule $\alpha_t$. The model then predicts the original tokens at the masked positions using the full context of $\rvx_t$.

Specifically, we apply Context-Adaptive Token-Level Noise Rescheduling~\citep{dream2025}:
\begin{equation}
    L(\theta) = -\E_{\rvx_0 \sim q(\rvx),t\sim \mathcal{U}(0,1), \rvx_t\sim q(\rvx_t|\rvx_0)} \sum_{n=1}^N \1_{[\mathrm{\rvx}_t^n=\texttt{MASK}]} w(t, \rvx_t, n) \log p_{\theta}(\rvx_0^n|\rvx_t),
\label{eq:loss_cart}
\end{equation}
where $w(t, \rvx_t, n)$ generalizes the term $w(t)$ and can be flexibly designed based on $\rvx_t$'s structure.
This approach assigns higher noise rates to harder-to-predict tokens, improving both convergence speed and final generation quality.

\subsection{Data Mixture}
We pretrain Dream-Coder 7B on a mixture of publicly available code, math, and general language datasets, totaling 322B tokens. The corpora include: (1) \textbf{OpenCoder}\footnote{\href{https://huggingface.co/collections/OpenCoder-LLM/opencoder-datasets-672e6db6a0fed24bd69ef1c2}{OpenCoder-LLM/opencoder-datasets}}~\citep{huang2025opencoderopencookbooktoptier}: Code-related data extracted from FineWeb~\citep{penedo2024fineweb}. (2) \textbf{Stack-Edu}\footnote{\href{https://huggingface.co/datasets/HuggingFaceTB/stack-edu}{HuggingFaceTB/stack-edu}}~\citep{allal2025smollm2smolgoesbig}: A dataset of educational code filtered from The Stack v2~\citep{lozhkov2024starcoder}, containing only the highest-quality educational programming content across 15 programming languages. (3) \textbf{Dolmino}\footnote{\href{https://huggingface.co/datasets/allenai/dolmino-mix-1124}{allenai/dolmino-mix-1124}}: A high-quality dataset mixture for annealing training. (4) \textbf{DCLM-Baseline}\footnote{\href{https://huggingface.co/datasets/mlfoundations/dclm-baseline-1.0}{mlfoundations/dclm-baseline-1.0}}~\citep{li2025dclm}: A dataset derived from CommonCrawl, providing high-quality general text to complement our code-focused training data.
These datasets collectively cover general programming tasks, algorithmic problem solving, software engineering best practices, and mathematical reasoning, ensuring broad coverage and robustness across coding tasks while maintaining strong general language capabilities.

\begin{figure}[t]
    \centering
    \begin{subfigure}{0.58\linewidth}
        \centering
        \includegraphics[width=\linewidth]{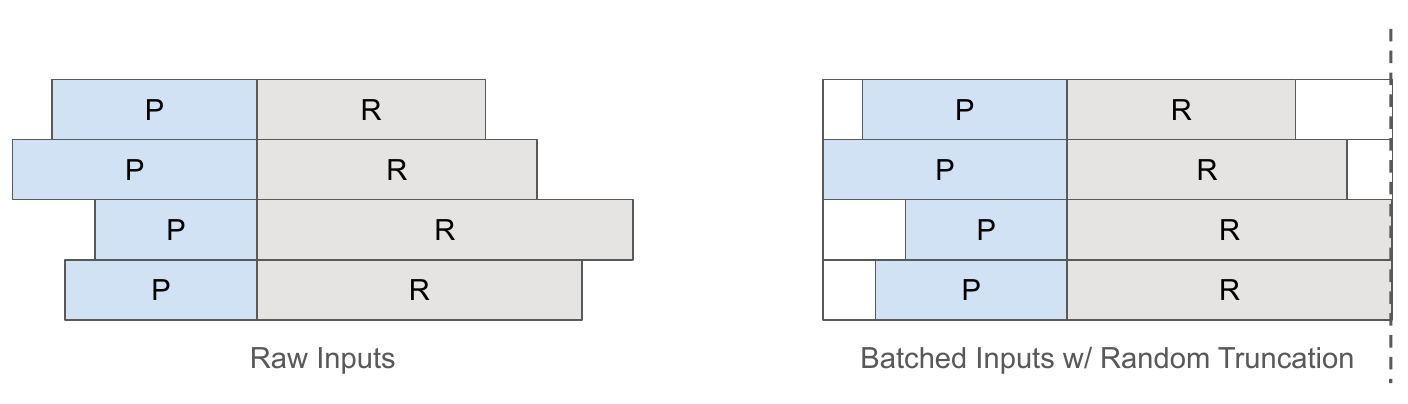}
        \caption{Illustration of the random truncation technique. \texttt{P}: prompt; \texttt{R}: response; white areas: \texttt{[PAD]} tokens.}
        \label{fig:truncation}
      \end{subfigure}%
      \hfill
      \begin{subfigure}{0.38\linewidth}
        \centering
        \includegraphics[width=\linewidth]{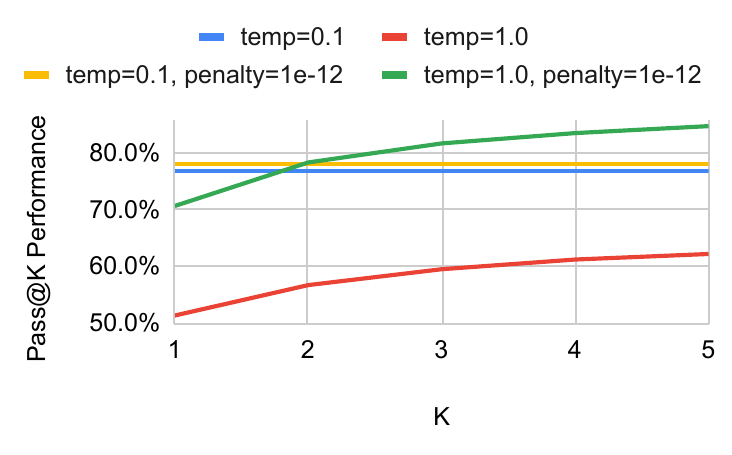}
        \caption{Pass@K analysis across different temperature and penalty configurations.}
        \label{fig:passk}
      \end{subfigure}%
\end{figure}

\section{Post-Training}
\label{sec:posttrain}

\subsection{Supervised Fine-Tuning}
We perform supervised fine-tuning on 5 million high-quality instruction-based code examples from Ling-Coder-SFT\footnote{\href{https://huggingface.co/datasets/inclusionAI/Ling-Coder-SFT}{inclusionAI/Ling-Coder-SFT}}~\citep{codefuse2025samplemattersleveragingmixtureofexperts} with rule-based filtering. During this phase, we observed two primary issues: low sample efficiency and generation instability due to excessive loss from padding tokens. To mitigate these:
\begin{itemize}
  \item \textbf{Random Truncation} (Figure~\ref{fig:truncation}): During training, we randomly truncate each response to the length of a randomly selected example in the batch, reducing wasted computation on \texttt{[PAD]} tokens and focusing learning on meaningful token predictions.
  \item \textbf{Padding Penalty}: We observe that the model tends to prematurely generate \texttt{[PAD]} tokens at high temperatures, degrading output quality. To counteract this behavior, we apply a gradually decaying penalty to the \texttt{[PAD]} token logits during inference, which discourages early sequence termination and provides better control over generation length.
\end{itemize}
This dual strategy significantly enhances sample utilization and stabilizes generation length without sacrificing accuracy (Figure~\ref{fig:passk}).

\subsection{Reinforcement Learning with Verifiable Rewards}
Building on the supervised model, we further conduct reinforcement learning with verifiable rewards to boost Dream-Coder's reasoning capabilities for code.

\paragraph{Data Curation.}
We assemble 17k prompts paired with unit-test-based rewards sourced from KodCode-V1\footnote{\href{https://huggingface.co/datasets/KodCode/KodCode-V1}{KodCode/KodCode-V1}}~\citep{xu2025kodcode}, DeepCoder-Preview\footnote{\href{https://huggingface.co/datasets/agentica-org/DeepCoder-Preview-Dataset}{agentica-org/DeepCoder-Preview-Dataset}}~\citep{deepcoder2025}, and Guru-RL-92k\footnote{\href{https://huggingface.co/datasets/LLM360/guru-RL-92k}{LLM360/guru-RL-92k}}~\citep{li2025codei,cheng2025revisiting}.
To ensure data quality and appropriate difficulty, we implement a three-phase filtering process: (1) \textbf{Quality filtering}: We retain only examples with at least 5 unit tests to ensure robust evaluation criteria; (2) \textbf{Deduplication}: We remove semantically similar prompts using a lightweight embedding model\footnote{\href{https://huggingface.co/sentence-transformers/all-MiniLM-L6-v2}{sentence-transformers/all-MiniLM-L6-v2}} to prevent data redundancy; and (3) \textbf{Difficulty calibration}: We filter out trivial problems by generating 8 candidate responses from Qwen2.5-Coder 7B Instruct and excluding prompts where all responses achieve perfect correctness, ensuring our dataset maintains an appropriate challenge level for reinforcement learning.

\paragraph{Training.} We apply reinforcement learning using the GRPO algorithm \citep{shao2024deepseekmath}, which maximizes:
\begin{equation*}
    \begin{gathered}
\mathbb{E}_{(q, a) \sim \mathcal{D},\left\{o_i\right\}_{i=1}^G \sim \pi_{\theta_{\mathrm{old}}}(\cdot \mid q),t\sim \mathcal{U}(0,1)}\left[\frac { 1 } { G } \sum _ { i = 1 } ^ { G } \frac { 1 } { | o _ { i } | } \sum _ { t = 1 } ^ { | o _ { i } | } \mathrm{min} \left(\rho_{i, t}(\theta) \hat{A}_{i},\mathrm{clip}\left(\rho_{i, t}(\theta), 1-\varepsilon_\mathrm{low}, 1+\varepsilon_\mathrm{high}\right) \hat{A}_{i}\right)\right],
\end{gathered}
\end{equation*}
where $\hat{A}_i=\frac{r\left(o_i\right)-\mathrm{mean}\left(\left\{r\left(o_j\right)\right\}_{j=1}^G\right)}{\mathrm{std}\left(\left\{r\left(o_j\right)\right\}_{j=1}^G\right)}$ and $\rho_{i, t}(\theta) = \frac{\pi_\theta\left(o_{i,t} \mid q, o_{i,t<T}\right)}{\pi_{\mathrm{old}}\left(o_{i,t} \mid q, o_{i,t<T}\right)}$.
Tailored for diffusion language models, we integrate several techniques to stabilize gradient estimation, inspired by recent work~\citep{yu2025dapo,gong2025diffucoder,Polaris2025}: (1) \textbf{No Entropy/KL Loss}: We omit entropy and KL divergence penalties to avoid training instability and encourage exploration. (2) \textbf{Clip-Higher}: We apply an asymmetric clipping range, extending the upper bound of importance weights ($\varepsilon_\mathrm{low}=0.2, \varepsilon_\mathrm{high}=0.28$) to explore high-reward tokens more aggressively. (3) \textbf{Coupled Sampling \& Informative Substitution}: For each batch, we sample complementary diffusion masks and perform intra-batch substitution of zero-advantage samples with high-advantage duplicates to maximize learning signal.
Additionally, we leverage Fast-dLLM~\citep{wu2025fastdllmtrainingfreeaccelerationdiffusion} for accelerated diffusion decoding. Code execution and reward evaluation are done in sandboxed environments~\citep{liu2024fullstack}.

\begin{figure}[t]
    \centering
    \includegraphics[width=0.8\linewidth]{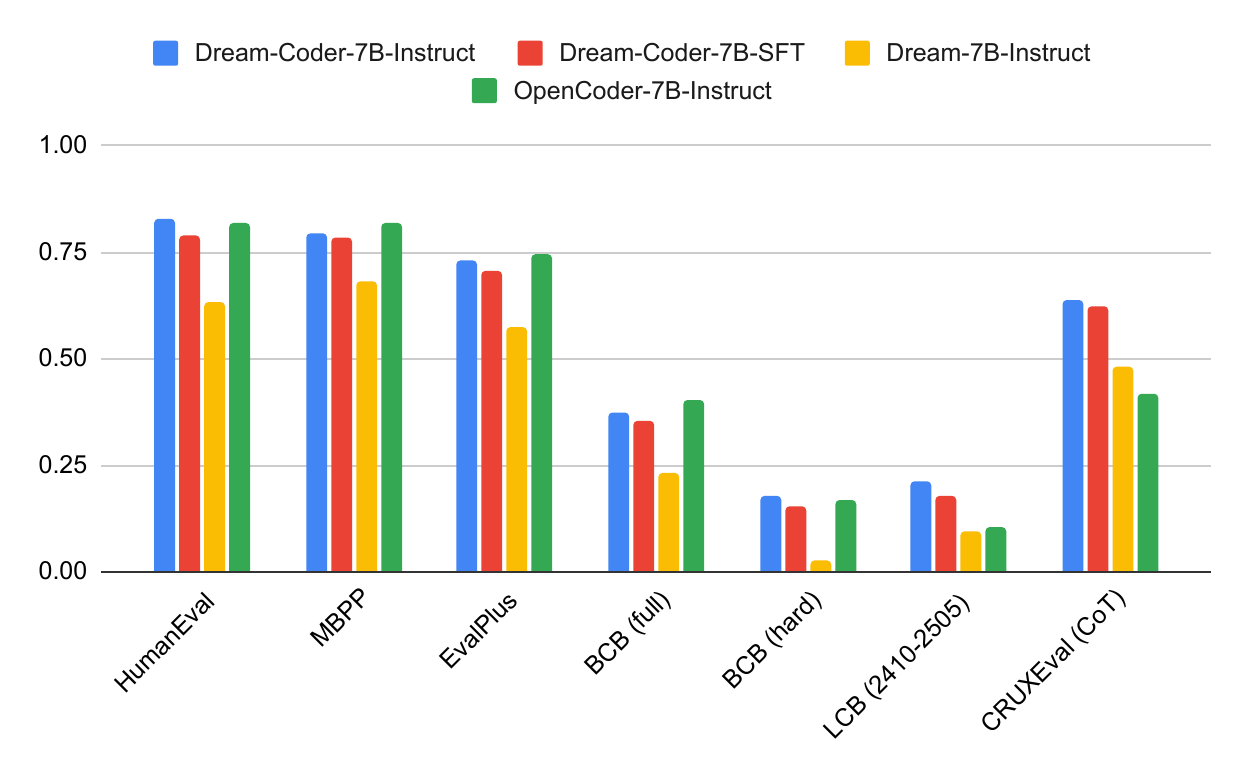}
    \caption{Overview of Dream-Coder 7B post-training pipeline showing supervised fine-tuning on Ling-Coder-SFT data followed by reinforcement learning with verifiable rewards using GRPO algorithm variants.}
    \label{fig:posttrain_pipeline}
\end{figure}

\section{Experiments}
\label{sec:experiments}

\subsection{Setup}
We evaluate Dream-Coder 7B Instruct on six standard code-generation benchmarks:
HumanEval~\citep{chen2021evaluating}, MBPP~\citep{austin2021program}, EvalPlus~\citep{liu2023your}, BigCodeBench~\citep{zhuo2024bigcodebench}, LiveCodeBench~\citep{jain2024livecodebench}, and CRUXEval~\citep{gu2024cruxeval}.
For general language understanding, we employ tasks such as MMLU~\citep{hendrycks2020measuring}, ARC-C~\citep{clark2018think}, HellaSwag~\citep{zellers2019hellaswag}, WinoGrande~\citep{sakaguchi2021winogrande}, and PIQA~\citep{bisk2020piqa}, and RACE~\citep{lai2017race} to test knowledge, reasoning, and commonsense abilities.
For mathematical and scientific reasoning, we use GSM8K~\citep{cobbe2021training} and MATH~\citep{hendrycks2020measuring} for numerical computation and problem-solving, and GPQA~\citep{rein2023gpqa} for scientific understanding.

\subsection{Evaluation on Base Models}
Table~\ref{tab:pretrain_code} shows the performance of Dream-Coder 7B on coding benchmarks, while Tables~\ref{tab:pretrain_general} and \ref{tab:pretrain_math} present its results on general benchmarks.

On coding benchmarks, Dream-Coder 7B demonstrates competitive performance, closely matching established autoregressive models like Qwen2.5-Coder-7B while significantly outperforming other diffusion-based models.
For general benchmarks, Dream-Coder 7B substantially outperforms baseline diffusion models while approaching the performance of state-of-the-art autoregressive models. The model shows robust performance across diverse tasks, indicating that the shift operation strategy successfully preserves general language understanding capabilities.

These results demonstrate that Dream-Coder 7B effectively bridges the performance gap between autoregressive and diffusion language models, achieving competitive coding performance while maintaining strong reasoning abilities across multiple domains.

\begin{table}[t!]
\centering
\scalebox{1.0}{
\footnotesize
\begin{tabular}{lcccccccc}
\toprule
 & \multicolumn{1}{c}{} & \multicolumn{2}{c}{\textbf{HumanEval}} & \multicolumn{2}{c}{\textbf{MBPP}} & \multicolumn{2}{c}{\textbf{BigCodeBench}} & \multicolumn{1}{c}{} \\
 & \multicolumn{1}{c}{\multirow{-2}{*}{\textbf{Open Data}}} & \multicolumn{1}{c}{{HE}} & \multicolumn{1}{c}{{HE+}} & \multicolumn{1}{c}{{MBPP}} & \multicolumn{1}{c}{{MBPP+}} & \multicolumn{1}{c}{{Full}} & \multicolumn{1}{c}{{Hard}} & \multicolumn{1}{c}{\multirow{-2}{*}{\textbf{Average}}} \\
 \midrule
\multicolumn{9}{l}{\textbf{Open-Weight AR Models}} \\
StarCoder2-7B & \cmark & 35.4 & 29.9 & 54.5 & 45.6 & 27.7 & 8.8 & 33.7 \\
DS-Coder-6.7B & \xmark & 47.6 & 39.6 & 70.2 & 56.6 & 41.1 & 11.5 & 44.4 \\
DS-Coder-V2-Lite & \xmark & 40.9 & 34.1 & 71.9 & 59.4 & 30.6 & 8.1 & 40.8 \\
OpenCoder-8B & \cmark & 66.5 & 63.4 & 79.9 & 70.4 & 40.5 & 9.5 & 55.0 \\
Seed-Coder-8B & \xmark & 77.4 & 68.3 & 82.0 & 69.0 & {-} & {-} & {-} \\
Qwen2.5-Coder-7B & \xmark & 61.6 & 53.0 & 75.9 & 62.9 & 45.8 & 16.2 & 52.6 \\
\midrule
\multicolumn{9}{l}{\textbf{Open-Weight Diffusion Models}} \\
LLaDA-8B & \xmark & 35.4 & 30.5 & 50.1 & 42.1 & 18.9 & 4.1 & 30.2 \\
DiffuCoder-7B & \cmark & \textbf{67.1} & 60.4 & 74.2 & 60.9 & \textbf{40.2} & 12.8 & 52.6 \\
Dream-7B & \cmark & 56.7 & 50.0 & 68.7 & 57.4 & 23.6 & 4.1 & 43.4 \\
\rowcolor{lightblue}
Dream-Coder-7B & \cmark & 66.5 & \textbf{60.4} & \textbf{75.9} & \textbf{61.6} & 38.5 & \textbf{14.2} & \textbf{52.9} \\
\bottomrule
\end{tabular}}
\caption{Base model performance comparison on coding benchmarks. We mark models trained on open-source data with {\cmark}, and those trained on in-house data with {\xmark}. For BigCodeBench, we report code completion scores. Our results are highlighted in \colorbox{lightblue}{blue}. The best results among open-weight diffusion language models are bolded.}
\label{tab:pretrain_code}
\end{table}

\begin{table}[t!]
\centering
\scalebox{0.93}{
\footnotesize
\begin{tabular}{lcccccccc}
\toprule
 & \textbf{Open Data} & \textbf{MMLU} & \textbf{ARC-C} & \textbf{Hellaswag} & \textbf{WinoGrande} & \textbf{PIQA} & \textbf{RACE} & \textbf{Average} \\
 \midrule
\multicolumn{7}{l}{\textbf{Open-Weight AR Models}} \\
OpenCoder-8B & \cmark & 40.0 & 35.3 & 56.7 & 61.1 & 71.9 & 36.3 & 50.2 \\
Seed-Coder-8B & \xmark & 41.7 & 34.8 & 52.4 & 56.1 & 69.4 & 34.2 & 48.1 \\
Qwen2.5-Coder-7B & \xmark & 68.1 & 48.6 & 75.6 & 72.3 & 79.6 & 37.8 & 63.7 \\
\midrule
\multicolumn{7}{l}{\textbf{Open-Weight Diffusion Models}} \\
DiffuCoder-7B & \cmark & 57.3 & 48.0 & 60.1 & 60.5 & 68.9 & 40.2 & 55.8 \\
\rowcolor{lightblue}
Dream-Coder-7B & \cmark & \textbf{65.6} & \textbf{53.8} & \textbf{67.3} & \textbf{66.2} & \textbf{73.3} & \textbf{43.4} & 61.6 \\
\bottomrule
\end{tabular}}
\caption{Performance of Dream-Coder 7B base model on general reasoning benchmarks compared to autoregressive and diffusion baselines.}
\label{tab:pretrain_general}
\end{table}

\begin{table}[]
\centering
\scalebox{1.0}{
\footnotesize
\begin{tabular}{lccccc}
\toprule
 & \textbf{Open Data} & \textbf{GSM8K} & \textbf{MATH} & \textbf{GPQA} & \textbf{Average} \\
 \midrule
\multicolumn{6}{l}{\textbf{Open-Weight AR Models}} \\
OpenCoder-8B & \cmark & 28.7 & 9.8 & 27.9 & 22.1 \\
Seed-Coder-8B & \xmark & 36.0 & 10.2 & 26.3 & 24.2 \\
Qwen2.5-Coder-7B & \xmark & 84.2 & 43.5 & 35.9 & 54.5 \\
\midrule
\multicolumn{6}{l}{\textbf{Open-Weight Diffusion Models}} \\
DiffuCoder-7B & \cmark & 51.8 & 17.3 & 25.7 & 31.6 \\
\rowcolor{lightblue}
Dream-Coder-7B & \cmark & \textbf{71.1} & \textbf{33.7} & \textbf{32.4} & \textbf{45.7} \\
\bottomrule
\end{tabular}}
\caption{Performance of Dream-Coder 7B base model on mathematical reasoning benchmarks (GSM8K and MATH) and science QA (GPQA) compared to baselines.}
\label{tab:pretrain_math}
\end{table}

\subsection{Evaluation on Instruct Models}
Tables~\ref{tab:instruct_code} and~\ref{tab:instruct_lcb_crux} report the performance of Dream-Coder 7B Instruct after post-training across various benchmarks.
On instruction-following coding tasks (Table~\ref{tab:instruct_code}), Dream-Coder 7B Instruct demonstrates strong performance, delivering results on par with top-tier models and outperforming other open-weight diffusion models by a wide margin.
On challenging real-world coding tasks, Dream-Coder 7B Instruct achieves \textbf{21.4\%} pass@1 on LiveCodeBench (Table~\ref{tab:instruct_lcb_crux})—on par with top proprietary systems (e.g.\ Mercury Coder Small at 22.9\%) and outperforming open-recipe models such as OpenCoder 8B Instruct.
These results confirm that diffusion language models can achieve competitive performance on real-world coding tasks when trained exclusively on public data, successfully bridging the gap between open-weight and proprietary systems while maintaining the architectural advantages of diffusion-based generation.

\begin{table}[t!]
\centering
\footnotesize
\begin{tabular}{lccccccc}
\toprule
                     &                                                                   &                                      &                                 &  &                                   & \multicolumn{2}{c}{\textbf{BigCodeBench}} \\
                     &  \multirow{-2}{*}{\textbf{Type}} & \multirow{-2}{*}{\textbf{Open Data}} & \multirow{-2}{*}{\textbf{HumanEval}} & \multirow{-2}{*}{\textbf{MBPP}} & \multirow{-2}{*}{\textbf{EvalPlus}} &  {Full}         &  {Hard}  \\
\midrule
\rowcolor[HTML]{E5E4E2}\multicolumn{8}{l}{\textbf{Proprietary Models}} \\
\rowcolor[HTML]{E5E4E2} Claude 3 Haiku                    & AR     & \xmark &  76.8 & 80.2 & 68.9 & 39.4 & 16.9 \\
\rowcolor[HTML]{E5E4E2} Mercury Coder Small                    & Diffusion     & \xmark &  90.0      & 80.4        & 76.4            & 45.3               & 22.3               \\
\rowcolor[HTML]{E5E4E2} Gemini Diffusion                       & Diffusion     & \xmark & 89.6      & 76.0        & -          & 45.4               & - \\
\midrule
\multicolumn{8}{l}{\textbf{Open-Weight AR Models}} \\
OpenCoder-8B-Instruct                          & AR            & \cmark & 81.7      & 82.0        & 74.4            & 40.3               & 16.9               \\
Qwen2.5-Coder-7B-Instruct                   & AR            & \xmark & 88.4      & 83.5        & 77.9            & 41.0               & 18.2               \\
Seed-Coder-8B-Instruct                         & AR            & \xmark & 84.8      & 85.2        & 75.0            & 44.5               & 26.4               \\
\midrule
\multicolumn{8}{l}{\textbf{Open-Weight Diffusion Models}} \\
LLaDA-1.5-8B                             & Diffusion     & \xmark & 52.4      & 42.8        & -          & - & - \\
Dream-7B-Instruct                              & Diffusion     & \cmark & 63.4      & 68.3        & 57.4            & 23.0 & 2.7                \\
DiffuCoder-7B-cpGRPO                            & Diffusion     & \cmark & 73.2      & 78.6        & 67.9            & \textbf{37.5}               & 10.8               \\
\rowcolor{lightblue} Dream-Coder-7B-Instruct                        & Diffusion     & \cmark & \textbf{82.9}      & \textbf{79.6}        & \textbf{73.1}            & 37.1               & \textbf{17.6}               \\
\bottomrule
\end{tabular}
\caption{Performance comparison of instruction-tuned models on HumanEval, MBPP, EvalPlus, and BigCodeBench.}
\label{tab:instruct_code}
\end{table}

\begin{table}[t!]
\centering
\footnotesize
\begin{tabular}{lcccccccc}
\toprule
                     &                                                                   &                                      & \multicolumn{4}{c}{\textbf{LiveCodeBench (2410-2505)}} & \multicolumn{2}{c}{\textbf{CRUXEval}}    \\
                     &  \multirow{-2}{*}{\textbf{Type}} & \multirow{-2}{*}{\textbf{Open Data}} &  Full & Easy & Medium & Hard     &  Input-CoT   &  {Output-CoT} \\
\midrule
\rowcolor[HTML]{E5E4E2}\multicolumn{9}{l}{\textbf{Proprietary Models}} \\
\rowcolor[HTML]{E5E4E2} Claude 3 Haiku                    & AR     & \xmark & 21.2 & 63.8 & 12.6 & 3.7 &  - & -                       \\
\rowcolor[HTML]{E5E4E2} Mercury Coder Small                    & Diffusion     & \xmark & 22.9 & 69.0 & 15.2 & 2.6 & 66.4 & 70.8 \\
\midrule
\multicolumn{9}{l}{\textbf{Open-Weight AR Models}} \\
OpenCoder-8B-Instruct                          & AR            & \cmark & 10.6 & 34.4 & 5.1 & 1.2 & 41.6 &	41.9 \\
Qwen2.5-Coder-7B-Instruct                   & AR            & \xmark & 19.3 & 62.7 & 7.4 & 3.4 & 65.8 &	65.9 \\
Seed-Coder-8B-Instruct                         & AR            & \xmark & 20.9 & 62.3 & 15.1 & 1.9 & 63.3 &	67.1 \\
\midrule
\multicolumn{9}{l}{\textbf{Open-Weight Diffusion Models}} \\
Dream-7B-Instruct                              & Diffusion     & \cmark &  9.7 & 34.5 &	1.9 &	1.3 &    53.1 &	42.8 \\
DiffuCoder-7B-cpGRPO                            & Diffusion     & \cmark & 10.3 & 35.7 & 2.9 & 1.3 & - & - \\
\rowcolor{lightblue} Dream-Coder-7B-Instruct                        & Diffusion     & \cmark & \textbf{21.4} & \textbf{64.3} & \textbf{12.4} & \textbf{3.9} & \textbf{64.6} &	\textbf{63.1}\\
\bottomrule
\end{tabular}
\caption{Performance comparison on LiveCodeBench (2410-2505) and CRUXEval benchmarks. Dream-Coder 7B Instruct achieves competitive results on challenging real-world coding tasks.}
\label{tab:instruct_lcb_crux}
\end{table}

\subsection{Generation Patterns}
Dream-Coder 7B exhibits three distinct generation patterns depending on the task complexity and structure, leveraging the flexibility of diffusion-based generation to adapt its approach dynamically. Unlike autoregressive models that are constrained to left-to-right token generation, our model demonstrates emergent adaptive behaviors that align with the inherent structure of coding tasks.

\begin{figure}[t!]
  \centering
  \begin{subfigure}{0.48\linewidth}
    \centering
    \includegraphics[width=\linewidth]{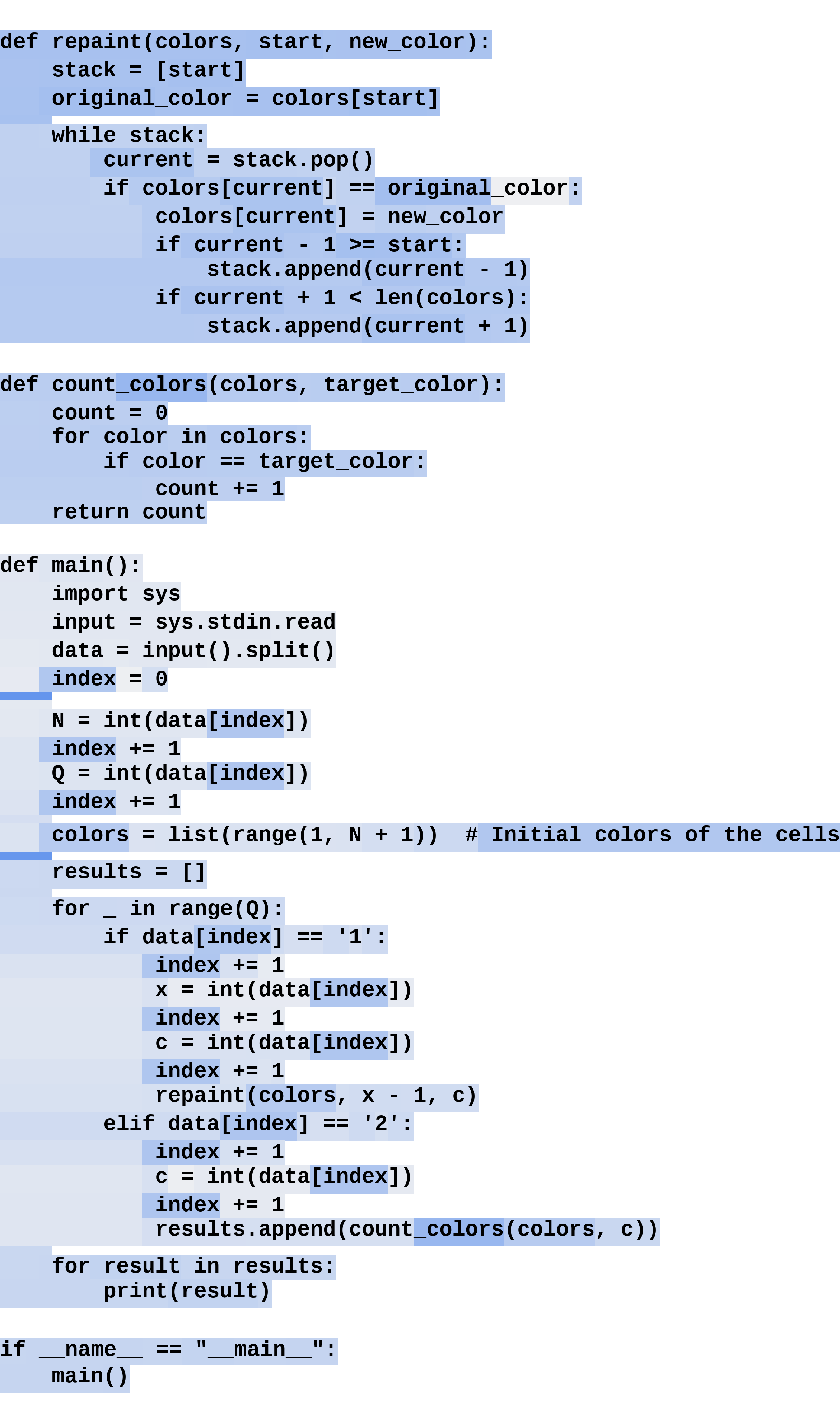}
    \caption{Sketch-First Generation (from LiveCodeBench)}
    \label{fig:lcb_patterns}
  \end{subfigure}%
  \hfill
  \begin{subfigure}{0.48\linewidth}
    \centering
    \includegraphics[width=\linewidth]{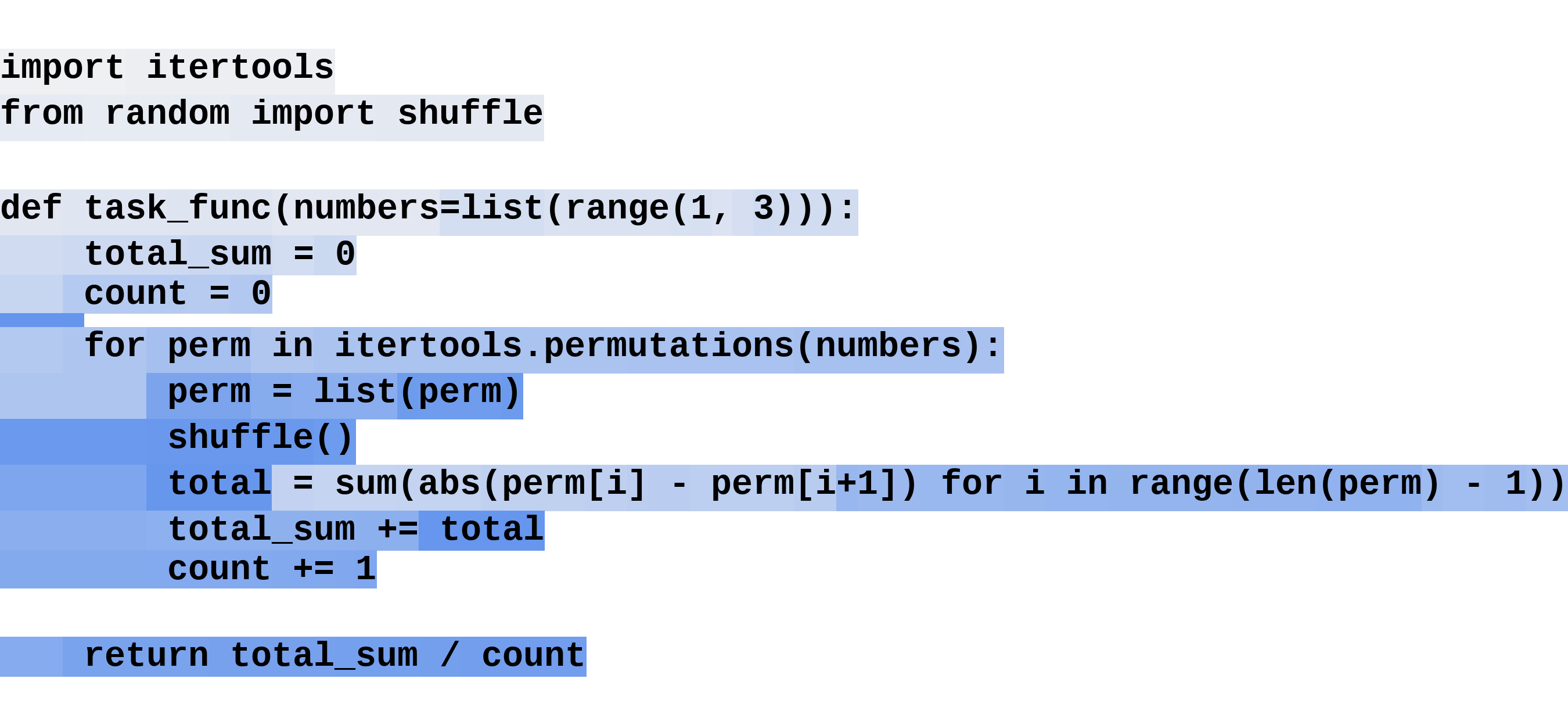}
    \caption{Left-to-Right Generation (from BigCodeBench)}
    \label{fig:bcb_patterns}
    
    \vspace{1.5cm}
    
    \includegraphics[width=\linewidth]{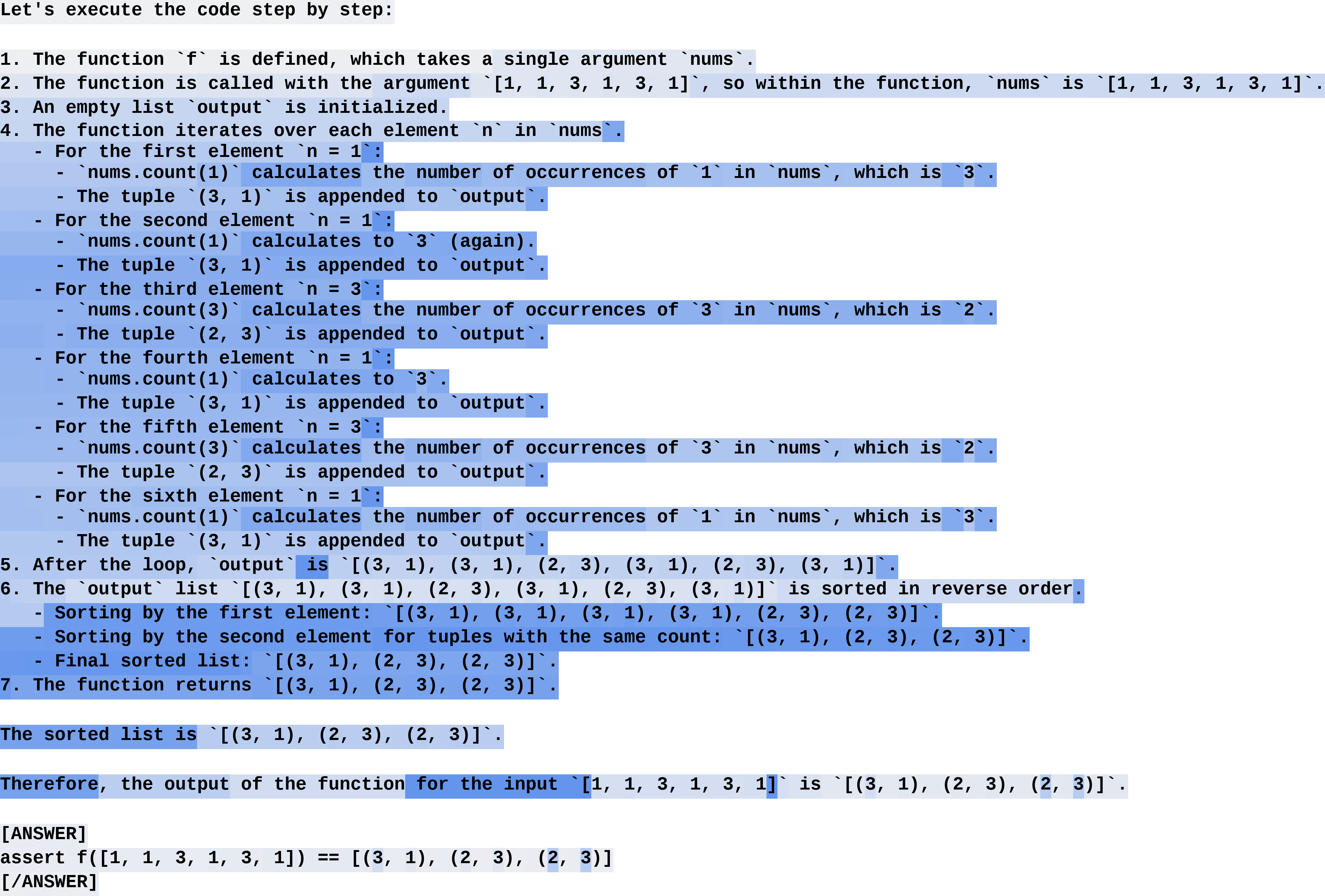}
    \caption{Interleaved Reasoning Generation (from CRUXEval)}
    \label{fig:cruxeval_patterns}
  \end{subfigure}
  
  \caption{Generation patterns exhibited by Dream-Coder 7B Instruct across different coding tasks. Colors encode the generation order during decoding (light to dark: first to last), revealing diverse non-autoregressive strategies.}
  \label{fig:generation_patterns}
\end{figure}

\paragraph{Sketch-First Scaffolding (Figure~\ref{fig:lcb_patterns}).} For template-heavy tasks requiring full program structure (e.g., LiveCodeBench), Dream-Coder 7B employs a sketch-first approach. The model first generates high-level structural elements such as function signatures, control flow constructs, and variable declarations, then progressively fills in implementation details. This pattern is particularly effective for complex algorithmic tasks where understanding the overall program flow is crucial before diving into specific logic. The diffusion process allows the model to establish this global structure early and refine it iteratively, leading to more coherent and correct programs.

\paragraph{Left-to-Right Completion(Figure~\ref{fig:bcb_patterns}).} For standalone function completions (e.g., HumanEval, MBPP), the model converges to a more traditional left-to-right pattern, but with enhanced flexibility. While the overall generation follows a sequential order, the diffusion mechanism enables local refinements and corrections that are impossible in pure autoregressive generation. This hybrid approach combines the efficiency of sequential generation with the error-correction capabilities of diffusion, resulting in higher success rates on standard completion benchmarks.

\paragraph{Interleaved Reasoning (Figure~\ref{fig:cruxeval_patterns})}. For logic-intensive tasks requiring multi-step reasoning (e.g., CRUXEval), Dream-Coder 7B demonstrates an interleaved pattern where logical components are generated in a non-linear fashion. The model may generate key logical conditions first, then fill in supporting code, and finally refine edge cases. This pattern reflects the human problem-solving approach where critical insights drive the overall solution structure, rather than purely sequential thinking.

\paragraph{Pattern Selection and Adaptation.} Importantly, these patterns are not pre-programmed but emerge naturally from the diffusion training process. The model learns to associate different generation strategies with different types of coding tasks through exposure to diverse training examples. During inference, the choice of pattern is influenced by the prompt structure, task complexity, and learned representations of similar problems encountered during training. This adaptive behavior represents a key advantage of diffusion-based generation over fixed autoregressive approaches.

The effectiveness of each pattern is validated by our benchmark results: sketch-first generation achieves superior performance on complex algorithmic tasks (LiveCodeBench), left-to-right completion excels on standard function completion (HumanEval/MBPP), and interleaved reasoning demonstrates strong results on logical reasoning tasks (CRUXEval). This pattern diversity contributes significantly to Dream-Coder 7B's competitive performance across diverse coding benchmarks.

\section{Related Work}
\label{sec:related}

\paragraph{Autoregressive Code Models.}
Autoregressive models have dominated code generation by generating tokens sequentially from left to right. Early specialized code models established strong baselines through large-scale pretraining on curated code repositories~\citep{roziere2023code,lozhkov2024starcoder,guo2024deepseek}. Recent efforts have explored enhancing coding capabilities through advanced reasoning~\citep{mcaleese2024llm,el2025competitive} and agentic frameworks that leverage tool use and multi-step planning~\citep{wei2025swe,team2025kimi}. However, autoregressive models are fundamentally limited to left-to-right generation, constraining their effectiveness in non-monotonic editing and scenarios requiring global planning or bidirectional context understanding~\citep{bachmann2024pitfalls,ye2024beyond}.

\paragraph{Diffusion Language Models.}
Diffusion models have emerged as a promising alternative to autoregressive generation, offering flexible generation orders and iterative refinement capabilities. 
\citet{austin2021structured, hoogeboom2021argmax} introduced discrete diffusion, laying the foundation for text diffusion modeling. Subsequent work further improves the text diffusion approaches from various perspectives~\citep{campbell2022continuous,li2022diffusion,gong2022diffuseq,zheng2023reparameterized,sahoo2024simple,shi2025simplifiedgeneralizedmaskeddiffusion,ou2024your}. Recent scaling efforts include SEDD~\citep{lou2023discrete}, Plaid~\citep{gulrajani2023likelihood}, LLaDA~\citep{nie2025llada}, and significant advances in adaptation techniques from pretrained autoregressive models, including DiffuLLaMA~\citep{gong2025scalingdiffusionlanguagemodels} and Dream~\citep{dream2025}. These models demonstrate competitive performance on language tasks while enabling novel capabilities like arbitrary-order generation and controllable text synthesis. Proprietary systems like Mercury~\citep{khanna2025mercury} and Gemini Diffusion~\citep{geminidiffusion} have shown competitive results through architectural optimizations and specialized training procedures. Recent work on accelerating diffusion inference addresses computational efficiency through KV caching and parallel generation strategies~\citep{ma2025dkv,wu2025fastdllmtrainingfreeaccelerationdiffusion,liu2025dllm,hu2025accelerating}, making diffusion models more practical for real-world applications.

\paragraph{Diffusion Models for Code Generation.}
Code generation often involves iterative refinements~\citep{xie2025teaching}, making discrete diffusion models that iteratively denoise from noise a natural choice for programming tasks. The application of diffusion models to code generation is an emerging area with growing interest. DiffuCoder~\citep{gong2025diffucoder} pioneered masked diffusion strategies specifically for programming languages, analyzing how diffusion objectives can capture code structure and dependencies. These advances highlight the potential for diffusion models to leverage the inherently iterative nature of programming workflows.

\section{Conclusion}
\label{sec:conclusion}
\base represents a continuation of our efforts to enhance open-source diffusion LLMs, with particular focus on post-training improvements and code generation capabilities. Trained entirely on open-source data, it delivers competitive performance with state-of-the-art autoregressive models while providing unique advantages for flexible code synthesis. Our experiments demonstrate that diffusion language models excel particularly in complex coding scenarios, where their non-autoregressive nature and iterative refinement process offer distinct benefits over traditional left-to-right generation approaches.

We are excited to further improve \base and \instruct. Future efforts will explore context extension, improved data curation strategies, and applications to specialized coding domains requiring advanced planning and reasoning abilities to further boost Dream models' capabilities.

\section*{Acknowledgments}
We thank Chengwu Cai for his support and helpful discussion. We also acknowledge the open-source community for providing high-quality datasets and evaluation frameworks. 
This research was supported in part by the joint research scheme of the National Natural Science Foundation of China (NSFC) and the Research Grants Council (RGC) under grant number
N\_HKU714/21.

\bibliographystyle{plainnat}
\bibliography{ref}

\begin{thebibliography}{65}
\providecommand{\natexlab}[1]{#1}
\providecommand{\url}[1]{\texttt{#1}}
\expandafter\ifx\csname urlstyle\endcsname\relax
  \providecommand{\doi}[1]{doi: #1}\else
  \providecommand{\doi}{doi: \begingroup \urlstyle{rm}\Url}\fi

\bibitem[Allal et~al.(2025)Allal, Lozhkov, Bakouch, Blázquez, Penedo, Tunstall, Marafioti, Kydlíček, Lajarín, Srivastav, Lochner, Fahlgren, Nguyen, Fourrier, Burtenshaw, Larcher, Zhao, Zakka, Morlon, Raffel, von Werra, and Wolf]{allal2025smollm2smolgoesbig}
Loubna~Ben Allal, Anton Lozhkov, Elie Bakouch, Gabriel~Martín Blázquez, Guilherme Penedo, Lewis Tunstall, Andrés Marafioti, Hynek Kydlíček, Agustín~Piqueres Lajarín, Vaibhav Srivastav, Joshua Lochner, Caleb Fahlgren, Xuan-Son Nguyen, Clémentine Fourrier, Ben Burtenshaw, Hugo Larcher, Haojun Zhao, Cyril Zakka, Mathieu Morlon, Colin Raffel, Leandro von Werra, and Thomas Wolf.
\newblock Smollm2: When smol goes big -- data-centric training of a small language model, 2025.
\newblock URL \url{https://arxiv.org/abs/2502.02737}.

\bibitem[An et~al.(2025)An, Xie, Li, Li, Zhang, Gong, Zhong, Xu, Qiu, Wang, and Kong]{Polaris2025}
Chenxin An, Zhihui Xie, Xiaonan Li, Lei Li, Jun Zhang, Shansan Gong, Ming Zhong, Jingjing Xu, Xipeng Qiu, Mingxuan Wang, and Lingpeng Kong.
\newblock Polaris: A post-training recipe for scaling reinforcement learning on advanced reasoning models.
\newblock 2025.
\newblock URL \url{https://hkunlp.github.io/blog/2025/Polaris}.

\bibitem[Anthropic(2025)]{claude_4}
Anthropic.
\newblock System card: Claude opus 4 \& claude sonnet 4, 2025.
\newblock URL \url{https://www-cdn.anthropic.com/07b2a3f9902ee19fe39a36ca638e5ae987bc64dd.pdf}.
\newblock System Card.

\bibitem[Austin et~al.(2021{\natexlab{a}})Austin, Johnson, Ho, Tarlow, and Van Den~Berg]{austin2021structured}
Jacob Austin, Daniel~D Johnson, Jonathan Ho, Daniel Tarlow, and Rianne Van Den~Berg.
\newblock Structured denoising diffusion models in discrete state-spaces.
\newblock \emph{Advances in neural information processing systems}, 34:\penalty0 17981--17993, 2021{\natexlab{a}}.

\bibitem[Austin et~al.(2021{\natexlab{b}})Austin, Odena, Nye, Bosma, Michalewski, Dohan, Jiang, Cai, Terry, Le, et~al.]{austin2021program}
Jacob Austin, Augustus Odena, Maxwell Nye, Maarten Bosma, Henryk Michalewski, David Dohan, Ellen Jiang, Carrie Cai, Michael Terry, Quoc Le, et~al.
\newblock Program synthesis with large language models.
\newblock \emph{arXiv preprint arXiv:2108.07732}, 2021{\natexlab{b}}.

\bibitem[Bachmann and Nagarajan(2024)]{bachmann2024pitfalls}
Gregor Bachmann and Vaishnavh Nagarajan.
\newblock The pitfalls of next-token prediction.
\newblock In \emph{International Conference on Machine Learning}, pages 2296--2318. PMLR, 2024.

\bibitem[Bisk et~al.(2020)Bisk, Zellers, Gao, Choi, et~al.]{bisk2020piqa}
Yonatan Bisk, Rowan Zellers, Jianfeng Gao, Yejin Choi, et~al.
\newblock Piqa: Reasoning about physical commonsense in natural language.
\newblock In \emph{Proceedings of the AAAI conference on artificial intelligence}, 2020.

\bibitem[Campbell et~al.(2022)Campbell, Benton, De~Bortoli, Rainforth, Deligiannidis, and Doucet]{campbell2022continuous}
Andrew Campbell, Joe Benton, Valentin De~Bortoli, Thomas Rainforth, George Deligiannidis, and Arnaud Doucet.
\newblock A continuous time framework for discrete denoising models.
\newblock \emph{Advances in Neural Information Processing Systems}, 35:\penalty0 28266--28279, 2022.

\bibitem[Chen et~al.(2021)Chen, Tworek, Jun, Yuan, Pinto, Kaplan, Edwards, Burda, Joseph, Brockman, et~al.]{chen2021evaluating}
Mark Chen, Jerry Tworek, Heewoo Jun, Qiming Yuan, Henrique Ponde De~Oliveira Pinto, Jared Kaplan, Harri Edwards, Yuri Burda, Nicholas Joseph, Greg Brockman, et~al.
\newblock Evaluating large language models trained on code.
\newblock \emph{arXiv preprint arXiv:2107.03374}, 2021.

\bibitem[Cheng et~al.(2025)Cheng, Hao, Liu, Zhou, Xie, Yao, Bian, Zhuang, Dey, Zha, Gu, Zhou, Wang, Li, Fan, She, Gao, Saparov, Li, Killian, Yurochkin, Liu, Xing, and Hu]{cheng2025revisiting}
Zhoujun Cheng, Shibo Hao, Tianyang Liu, Fan Zhou, Yutao Xie, Feng Yao, Yuexin Bian, Yonghao Zhuang, Nilabjo Dey, Yuheng Zha, Yi~Gu, Kun Zhou, Yuqi Wang, Yuan Li, Richard Fan, Jianshu She, Chengqian Gao, Abulhair Saparov, Haonan Li, Taylor~W. Killian, Mikhail Yurochkin, Zhengzhong Liu, Eric~P. Xing, and Zhiting Hu.
\newblock Revisiting reinforcement learning for llm reasoning from a cross-domain perspective, 2025.
\newblock URL \url{https://arxiv.org/abs/2506.14965}.

\bibitem[Clark et~al.(2018)Clark, Cowhey, Etzioni, Khot, Sabharwal, Schoenick, and Tafjord]{clark2018think}
Peter Clark, Isaac Cowhey, Oren Etzioni, Tushar Khot, Ashish Sabharwal, Carissa Schoenick, and Oyvind Tafjord.
\newblock Think you have solved question answering? try arc, the ai2 reasoning challenge.
\newblock \emph{arXiv preprint arXiv:1803.05457}, 2018.

\bibitem[Cobbe et~al.(2021)Cobbe, Kosaraju, Bavarian, Chen, Jun, Kaiser, Plappert, Tworek, Hilton, Nakano, et~al.]{cobbe2021training}
Karl Cobbe, Vineet Kosaraju, Mohammad Bavarian, Mark Chen, Heewoo Jun, Lukasz Kaiser, Matthias Plappert, Jerry Tworek, Jacob Hilton, Reiichiro Nakano, et~al.
\newblock Training verifiers to solve math word problems.
\newblock \emph{arXiv preprint arXiv:2110.14168}, 2021.

\bibitem[Codefuse and Team(2025)]{codefuse2025samplemattersleveragingmixtureofexperts}
Codefuse and Ling Team.
\newblock Every sample matters: Leveraging mixture-of-experts and high-quality data for efficient and accurate code llm, 2025.
\newblock URL \url{https://arxiv.org/abs/2503.17793}.

\bibitem[Comanici et~al.(2025)Comanici, Bieber, Schaekermann, Pasupat, Sachdeva, Dhillon, Blistein, Ram, Zhang, Rosen, et~al.]{comanici2025gemini}
Gheorghe Comanici, Eric Bieber, Mike Schaekermann, Ice Pasupat, Noveen Sachdeva, Inderjit Dhillon, Marcel Blistein, Ori Ram, Dan Zhang, Evan Rosen, et~al.
\newblock Gemini 2.5: Pushing the frontier with advanced reasoning, multimodality, long context, and next generation agentic capabilities.
\newblock \emph{arXiv preprint arXiv:2507.06261}, 2025.

\bibitem[DeepMind(2025)]{geminidiffusion}
DeepMind.
\newblock Gemini diffusion.
\newblock 2025.
\newblock URL \url{https://deepmind.google/models/gemini-diffusion/}.

\bibitem[El-Kishky et~al.(2025)El-Kishky, Wei, Saraiva, Minaiev, Selsam, Dohan, Song, Lightman, Clavera, Pachocki, et~al.]{el2025competitive}
Ahmed El-Kishky, Alexander Wei, Andre Saraiva, Borys Minaiev, Daniel Selsam, David Dohan, Francis Song, Hunter Lightman, Ignasi Clavera, Jakub Pachocki, et~al.
\newblock Competitive programming with large reasoning models.
\newblock \emph{arXiv preprint arXiv:2502.06807}, 2025.

\bibitem[Gong et~al.(2023)Gong, Li, Feng, Wu, and Kong]{gong2022diffuseq}
Shansan Gong, Mukai Li, Jiangtao Feng, Zhiyong Wu, and Lingpeng Kong.
\newblock {DiffuSeq}: Sequence to sequence text generation with diffusion models.
\newblock In \emph{International Conference on Learning Representations, ICLR}, 2023.

\bibitem[Gong et~al.(2025{\natexlab{a}})Gong, Agarwal, Zhang, Ye, Zheng, Li, An, Zhao, Bi, Han, Peng, and Kong]{gong2025scalingdiffusionlanguagemodels}
Shansan Gong, Shivam Agarwal, Yizhe Zhang, Jiacheng Ye, Lin Zheng, Mukai Li, Chenxin An, Peilin Zhao, Wei Bi, Jiawei Han, Hao Peng, and Lingpeng Kong.
\newblock Scaling diffusion language models via adaptation from autoregressive models.
\newblock \emph{International Conference on Learning Representations}, 2025{\natexlab{a}}.

\bibitem[Gong et~al.(2025{\natexlab{b}})Gong, Zhang, Zheng, Gu, Jaitly, Kong, and Zhang]{gong2025diffucoder}
Shansan Gong, Ruixiang Zhang, Huangjie Zheng, Jiatao Gu, Navdeep Jaitly, Lingpeng Kong, and Yizhe Zhang.
\newblock Diffucoder: Understanding and improving masked diffusion models for code generation.
\newblock \emph{arXiv preprint arXiv:2506.20639}, 2025{\natexlab{b}}.

\bibitem[Gu et~al.(2024)Gu, Rozi{\`e}re, Leather, Solar-Lezama, Synnaeve, and Wang]{gu2024cruxeval}
Alex Gu, Baptiste Rozi{\`e}re, Hugh Leather, Armando Solar-Lezama, Gabriel Synnaeve, and Sida~I Wang.
\newblock Cruxeval: A benchmark for code reasoning, understanding and execution.
\newblock \emph{arXiv preprint arXiv:2401.03065}, 2024.

\bibitem[Gulrajani and Hashimoto(2023)]{gulrajani2023likelihood}
Ishaan Gulrajani and Tatsunori~B Hashimoto.
\newblock Likelihood-based diffusion language models.
\newblock \emph{Advances in Neural Information Processing Systems}, 36:\penalty0 16693--16715, 2023.

\bibitem[Guo et~al.(2024)Guo, Zhu, Yang, Xie, Dong, Zhang, Chen, Bi, Wu, Li, et~al.]{guo2024deepseek}
Daya Guo, Qihao Zhu, Dejian Yang, Zhenda Xie, Kai Dong, Wentao Zhang, Guanting Chen, Xiao Bi, Yu~Wu, YK~Li, et~al.
\newblock Deepseek-coder: When the large language model meets programming--the rise of code intelligence.
\newblock \emph{arXiv preprint arXiv:2401.14196}, 2024.

\bibitem[Hendrycks et~al.(2020)Hendrycks, Burns, Basart, Zou, Mazeika, Song, and Steinhardt]{hendrycks2020measuring}
Dan Hendrycks, Collin Burns, Steven Basart, Andy Zou, Mantas Mazeika, Dawn Song, and Jacob Steinhardt.
\newblock Measuring massive multitask language understanding.
\newblock \emph{arXiv preprint arXiv:2009.03300}, 2020.

\bibitem[Hoogeboom et~al.(2021)Hoogeboom, Nielsen, Jaini, Forr{\'e}, and Welling]{hoogeboom2021argmax}
Emiel Hoogeboom, Didrik Nielsen, Priyank Jaini, Patrick Forr{\'e}, and Max Welling.
\newblock Argmax flows and multinomial diffusion: Learning categorical distributions.
\newblock \emph{Advances in neural information processing systems}, 34:\penalty0 12454--12465, 2021.

\bibitem[Hu et~al.(2025)Hu, Meng, Akhauri, Abdelfattah, Seo, Zhang, and Gupta]{hu2025accelerating}
Zhanqiu Hu, Jian Meng, Yash Akhauri, Mohamed~S Abdelfattah, Jae-sun Seo, Zhiru Zhang, and Udit Gupta.
\newblock Accelerating diffusion language model inference via efficient kv caching and guided diffusion.
\newblock \emph{arXiv preprint arXiv:2505.21467}, 2025.

\bibitem[Huang et~al.(2025)Huang, Cheng, Liu, Hao, Song, Xu, Yang, Liu, Zhang, Chai, Yuan, Zhang, Fu, Liu, Zhang, Wang, Qi, Xu, and Chu]{huang2025opencoderopencookbooktoptier}
Siming Huang, Tianhao Cheng, J.~K. Liu, Jiaran Hao, Liuyihan Song, Yang Xu, J.~Yang, Jiaheng Liu, Chenchen Zhang, Linzheng Chai, Ruifeng Yuan, Zhaoxiang Zhang, Jie Fu, Qian Liu, Ge~Zhang, Zili Wang, Yuan Qi, Yinghui Xu, and Wei Chu.
\newblock Opencoder: The open cookbook for top-tier code large language models, 2025.
\newblock URL \url{https://arxiv.org/abs/2411.04905}.

\bibitem[Hui et~al.(2024)Hui, Yang, Cui, Yang, Liu, Zhang, Liu, Zhang, Yu, Lu, et~al.]{hui2024qwen2}
Binyuan Hui, Jian Yang, Zeyu Cui, Jiaxi Yang, Dayiheng Liu, Lei Zhang, Tianyu Liu, Jiajun Zhang, Bowen Yu, Keming Lu, et~al.
\newblock Qwen2. 5-coder technical report.
\newblock \emph{arXiv preprint arXiv:2409.12186}, 2024.

\bibitem[Jain et~al.(2024)Jain, Han, Gu, Li, Yan, Zhang, Wang, Solar-Lezama, Sen, and Stoica]{jain2024livecodebench}
Naman Jain, King Han, Alex Gu, Wen-Ding Li, Fanjia Yan, Tianjun Zhang, Sida Wang, Armando Solar-Lezama, Koushik Sen, and Ion Stoica.
\newblock Livecodebench: Holistic and contamination free evaluation of large language models for code.
\newblock \emph{arXiv preprint arXiv:2403.07974}, 2024.

\bibitem[Khanna et~al.(2025)Khanna, Kharbanda, Li, Varma, Wang, Birnbaum, Luo, Miraoui, Palrecha, Ermon, et~al.]{khanna2025mercury}
Samar Khanna, Siddhant Kharbanda, Shufan Li, Harshit Varma, Eric Wang, Sawyer Birnbaum, Ziyang Luo, Yanis Miraoui, Akash Palrecha, Stefano Ermon, et~al.
\newblock Mercury: Ultra-fast language models based on diffusion.
\newblock \emph{arXiv e-prints}, pages arXiv--2506, 2025.

\bibitem[Kingma et~al.(2021)Kingma, Salimans, Poole, and Ho]{kingma2021variational}
Diederik Kingma, Tim Salimans, Ben Poole, and Jonathan Ho.
\newblock Variational diffusion models.
\newblock \emph{Advances in neural information processing systems}, 34:\penalty0 21696--21707, 2021.

\bibitem[Lai et~al.(2017)Lai, Xie, Liu, Yang, and Hovy]{lai2017race}
Guokun Lai, Qizhe Xie, Hanxiao Liu, Yiming Yang, and Eduard Hovy.
\newblock Race: Large-scale reading comprehension dataset from examinations.
\newblock \emph{arXiv preprint arXiv:1704.04683}, 2017.

\bibitem[Li et~al.(2024)Li, Fang, Smyrnis, Ivgi, Jordan, Gadre, Bansal, Guha, Keh, Arora, Garg, Xin, Muennighoff, Heckel, Mercat, Chen, Gururangan, Wortsman, Albalak, Bitton, Nezhurina, Abbas, Hsieh, Ghosh, Gardner, Kilian, Zhang, Shao, Pratt, Sanyal, Ilharco, Daras, Marathe, Gokaslan, Zhang, Chandu, Nguyen, Vasiljevic, Kakade, Song, Sanghavi, Faghri, Oh, Zettlemoyer, Lo, El-Nouby, Pouransari, Toshev, Wang, Groeneveld, Soldaini, Koh, Jitsev, Kollar, Dimakis, Carmon, Dave, Schmidt, and Shankar]{li2025dclm}
Jeffrey Li, Alex Fang, Georgios Smyrnis, Maor Ivgi, Matt Jordan, Samir Gadre, Hritik Bansal, Etash Guha, Sedrick Keh, Kushal Arora, Saurabh Garg, Rui Xin, Niklas Muennighoff, Reinhard Heckel, Jean Mercat, Mayee Chen, Suchin Gururangan, Mitchell Wortsman, Alon Albalak, Yonatan Bitton, Marianna Nezhurina, Amro Abbas, Cheng-Yu Hsieh, Dhruba Ghosh, Josh Gardner, Maciej Kilian, Hanlin Zhang, Rulin Shao, Sarah Pratt, Sunny Sanyal, Gabriel Ilharco, Giannis Daras, Kalyani Marathe, Aaron Gokaslan, Jieyu Zhang, Khyathi Chandu, Thao Nguyen, Igor Vasiljevic, Sham Kakade, Shuran Song, Sujay Sanghavi, Fartash Faghri, Sewoong Oh, Luke Zettlemoyer, Kyle Lo, Alaaeldin El-Nouby, Hadi Pouransari, Alexander Toshev, Stephanie Wang, Dirk Groeneveld, Luca Soldaini, Pang~Wei Koh, Jenia Jitsev, Thomas Kollar, Alexandros~G. Dimakis, Yair Carmon, Achal Dave, Ludwig Schmidt, and Vaishaal Shankar.
\newblock Datacomp-lm: In search of the next generation of training sets for language models.
\newblock \emph{Advances in neural information processing systems}, 2024.

\bibitem[Li et~al.(2025)Li, Guo, Yang, Xu, Wu, and He]{li2025codei}
Junlong Li, Daya Guo, Dejian Yang, Runxin Xu, Yu~Wu, and Junxian He.
\newblock Codei/o: Condensing reasoning patterns via code input-output prediction.
\newblock \emph{arXiv preprint arXiv:2502.07316}, 2025.

\bibitem[Li et~al.(2022)Li, Thickstun, Gulrajani, Liang, and Hashimoto]{li2022diffusion}
Xiang~Lisa Li, John Thickstun, Ishaan Gulrajani, Percy Liang, and Tatsunori~B Hashimoto.
\newblock Diffusion-lm improves controllable text generation.
\newblock In \emph{Conference on Neural Information Processing Systems, NeurIPS}, 2022.

\bibitem[Liu et~al.(2023)Liu, Xia, Wang, and Zhang]{liu2023your}
Jiawei Liu, Chunqiu~Steven Xia, Yuyao Wang, and Lingming Zhang.
\newblock Is your code generated by chatgpt really correct? rigorous evaluation of large language models for code generation.
\newblock \emph{Advances in Neural Information Processing Systems}, 36:\penalty0 21558--21572, 2023.

\bibitem[Liu et~al.(2024)Liu, Zhu, Liu, Xin, Li, Long, Chen, Yang, Xia, Peng, et~al.]{liu2024fullstack}
Siyao Liu, He~Zhu, Jerry Liu, Shulin Xin, Aoyan Li, Rui Long, Li~Chen, Jack Yang, Jinxiang Xia, ZY~Peng, et~al.
\newblock Fullstack bench: Evaluating llms as full stack coder.
\newblock \emph{arXiv preprint arXiv:2412.00535}, 2024.

\bibitem[Liu et~al.(2025)Liu, Yang, Zhang, Chen, Zou, Wei, Wang, and Zhang]{liu2025dllm}
Zhiyuan Liu, Yicun Yang, Yaojie Zhang, Junjie Chen, Chang Zou, Qingyuan Wei, Shaobo Wang, and Linfeng Zhang.
\newblock dllm-cache: Accelerating diffusion large language models with adaptive caching.
\newblock \emph{arXiv preprint arXiv:2506.06295}, 2025.

\bibitem[Lou et~al.(2024)Lou, Meng, and Ermon]{lou2023discrete}
Aaron Lou, Chenlin Meng, and Stefano Ermon.
\newblock Discrete diffusion language modeling by estimating the ratios of the data distribution.
\newblock In \emph{International Conference on Machine Learning, ICML}, 2024.

\bibitem[Lozhkov et~al.(2024)Lozhkov, Li, Allal, Cassano, Lamy-Poirier, Tazi, Tang, Pykhtar, Liu, Wei, Liu, Tian, Kocetkov, Zucker, Belkada, Wang, Liu, Abulkhanov, Paul, Li, Li, Risdal, Li, Zhu, Zhuo, Zheltonozhskii, Dade, Yu, Krauß, Jain, Su, He, Dey, Abati, Chai, Muennighoff, Tang, Oblokulov, Akiki, Marone, Mou, Mishra, Gu, Hui, Dao, Zebaze, Dehaene, Patry, Xu, McAuley, Hu, Scholak, Paquet, Robinson, Anderson, Chapados, Patwary, Tajbakhsh, Jernite, Ferrandis, Zhang, Hughes, Wolf, Guha, von Werra, and de~Vries]{lozhkov2024starcoder}
Anton Lozhkov, Raymond Li, Loubna~Ben Allal, Federico Cassano, Joel Lamy-Poirier, Nouamane Tazi, Ao~Tang, Dmytro Pykhtar, Jiawei Liu, Yuxiang Wei, Tianyang Liu, Max Tian, Denis Kocetkov, Arthur Zucker, Younes Belkada, Zijian Wang, Qian Liu, Dmitry Abulkhanov, Indraneil Paul, Zhuang Li, Wen-Ding Li, Megan Risdal, Jia Li, Jian Zhu, Terry~Yue Zhuo, Evgenii Zheltonozhskii, Nii Osae~Osae Dade, Wenhao Yu, Lucas Krauß, Naman Jain, Yixuan Su, Xuanli He, Manan Dey, Edoardo Abati, Yekun Chai, Niklas Muennighoff, Xiangru Tang, Muhtasham Oblokulov, Christopher Akiki, Marc Marone, Chenghao Mou, Mayank Mishra, Alex Gu, Binyuan Hui, Tri Dao, Armel Zebaze, Olivier Dehaene, Nicolas Patry, Canwen Xu, Julian McAuley, Han Hu, Torsten Scholak, Sebastien Paquet, Jennifer Robinson, Carolyn~Jane Anderson, Nicolas Chapados, Mostofa Patwary, Nima Tajbakhsh, Yacine Jernite, Carlos~Muñoz Ferrandis, Lingming Zhang, Sean Hughes, Thomas Wolf, Arjun Guha, Leandro von Werra, and Harm de~Vries.
\newblock Starcoder 2 and the stack v2: The next generation, 2024.

\bibitem[Luo et~al.(2025)Luo, Tan, Huang, Patel, Ariyak, Wu, Shi, Xin, Cai, Weber, Zhang, Li, Popa, and Stoica]{deepcoder2025}
Michael Luo, Sijun Tan, Roy Huang, Ameen Patel, Alpay Ariyak, Qingyang Wu, Xiaoxiang Shi, Rachel Xin, Colin Cai, Maurice Weber, Ce~Zhang, Li~Erran Li, Raluca~Ada Popa, and Ion Stoica.
\newblock Deepcoder: A fully open-source 14b coder at o3-mini level.
\newblock \url{https://pretty-radio-b75.notion.site/DeepCoder-A-Fully-Open-Source-14B-Coder-at-O3-mini-Level-1cf81902c14680b3bee5eb349a512a51}, 2025.
\newblock Notion Blog.

\bibitem[Ma et~al.(2025)Ma, Yu, Fang, and Wang]{ma2025dkv}
Xinyin Ma, Runpeng Yu, Gongfan Fang, and Xinchao Wang.
\newblock dkv-cache: The cache for diffusion language models.
\newblock \emph{arXiv preprint arXiv:2505.15781}, 2025.

\bibitem[McAleese et~al.(2024)McAleese, Pokorny, Uribe, Nitishinskaya, Trebacz, and Leike]{mcaleese2024llm}
Nat McAleese, Rai~Michael Pokorny, Juan Felipe~Ceron Uribe, Evgenia Nitishinskaya, Maja Trebacz, and Jan Leike.
\newblock Llm critics help catch llm bugs.
\newblock \emph{arXiv preprint arXiv:2407.00215}, 2024.

\bibitem[Nie et~al.(2025{\natexlab{a}})Nie, Zhu, You, Zhang, Ou, Hu, Zhou, Lin, Wen, and Li]{nie2025large}
Shen Nie, Fengqi Zhu, Zebin You, Xiaolu Zhang, Jingyang Ou, Jun Hu, Jun Zhou, Yankai Lin, Ji-Rong Wen, and Chongxuan Li.
\newblock Large language diffusion models.
\newblock \emph{arXiv preprint arXiv:2502.09992}, 2025{\natexlab{a}}.

\bibitem[Nie et~al.(2025{\natexlab{b}})Nie, Zhu, You, Zhang, Ou, Hu, Zhou, Lin, Wen, and Li]{nie2025llada}
Shen Nie, Fengqi Zhu, Zebin You, Xiaolu Zhang, Jingyang Ou, Jun Hu, Jun Zhou, Yankai Lin, Ji-Rong Wen, and Chongxuan Li.
\newblock Large language diffusion models.
\newblock \emph{arXiv preprint arXiv:2502.09992}, 2025{\natexlab{b}}.

\bibitem[OpenAI(2025)]{openAI_o3_o4_mini}
OpenAI.
\newblock Openai o3 and o4-mini system card, 2025.
\newblock URL \url{https://cdn.openai.com/pdf/2221c875-02dc-4789-800b-e7758f3722c1/o3-and-o4-mini-system-card.pdf}.
\newblock System Card.

\bibitem[Ou et~al.(2025)Ou, Nie, Xue, Zhu, Sun, Li, and Li]{ou2024your}
Jingyang Ou, Shen Nie, Kaiwen Xue, Fengqi Zhu, Jiacheng Sun, Zhenguo Li, and Chongxuan Li.
\newblock Your absorbing discrete diffusion secretly models the conditional distributions of clean data.
\newblock \emph{International Conference on Learning Representations}, 2025.

\bibitem[Penedo et~al.(2024)Penedo, Kydl{\'\i}{\v{c}}ek, Lozhkov, Mitchell, Raffel, Von~Werra, Wolf, et~al.]{penedo2024fineweb}
Guilherme Penedo, Hynek Kydl{\'\i}{\v{c}}ek, Anton Lozhkov, Margaret Mitchell, Colin~A Raffel, Leandro Von~Werra, Thomas Wolf, et~al.
\newblock The fineweb datasets: Decanting the web for the finest text data at scale.
\newblock \emph{Advances in Neural Information Processing Systems}, 37:\penalty0 30811--30849, 2024.

\bibitem[Rein et~al.(2023)Rein, Hou, Stickland, Petty, Pang, Dirani, Michael, and Bowman]{rein2023gpqa}
David Rein, Betty~Li Hou, Asa~Cooper Stickland, Jackson Petty, Richard~Yuanzhe Pang, Julien Dirani, Julian Michael, and Samuel~R Bowman.
\newblock Gpqa: A graduate-level google-proof q\&a benchmark.
\newblock \emph{arXiv preprint arXiv:2311.12022}, 2023.

\bibitem[Roziere et~al.(2023)Roziere, Gehring, Gloeckle, Sootla, Gat, Tan, Adi, Liu, Sauvestre, Remez, et~al.]{roziere2023code}
Baptiste Roziere, Jonas Gehring, Fabian Gloeckle, Sten Sootla, Itai Gat, Xiaoqing~Ellen Tan, Yossi Adi, Jingyu Liu, Romain Sauvestre, Tal Remez, et~al.
\newblock Code llama: Open foundation models for code.
\newblock \emph{arXiv preprint arXiv:2308.12950}, 2023.

\bibitem[Sahoo et~al.(2024)Sahoo, Arriola, Gokaslan, Marroquin, Rush, Schiff, Chiu, and Kuleshov]{sahoo2024simple}
Subham~Sekhar Sahoo, Marianne Arriola, Aaron Gokaslan, Edgar~Mariano Marroquin, Alexander~M Rush, Yair Schiff, Justin~T Chiu, and Volodymyr Kuleshov.
\newblock Simple and effective masked diffusion language models.
\newblock In \emph{The Thirty-eighth Annual Conference on Neural Information Processing Systems}, 2024.
\newblock URL \url{https://openreview.net/forum?id=L4uaAR4ArM}.

\bibitem[Sakaguchi et~al.(2021)Sakaguchi, Bras, Bhagavatula, and Choi]{sakaguchi2021winogrande}
Keisuke Sakaguchi, Ronan~Le Bras, Chandra Bhagavatula, and Yejin Choi.
\newblock Winogrande: An adversarial winograd schema challenge at scale.
\newblock \emph{Communications of the ACM}, 64\penalty0 (9):\penalty0 99--106, 2021.

\bibitem[Shao et~al.(2024)Shao, Wang, Zhu, Xu, Song, Bi, Zhang, Zhang, Li, Wu, et~al.]{shao2024deepseekmath}
Zhihong Shao, Peiyi Wang, Qihao Zhu, Runxin Xu, Junxiao Song, Xiao Bi, Haowei Zhang, Mingchuan Zhang, YK~Li, Yang Wu, et~al.
\newblock Deepseekmath: Pushing the limits of mathematical reasoning in open language models.
\newblock \emph{arXiv preprint arXiv:2402.03300}, 2024.

\bibitem[Shi et~al.(2024)Shi, Han, Wang, Doucet, and Titsias]{shi2025simplifiedgeneralizedmaskeddiffusion}
Jiaxin Shi, Kehang Han, Zhe Wang, Arnaud Doucet, and Michalis~K. Titsias.
\newblock Simplified and generalized masked diffusion for discrete data.
\newblock \emph{Neural Information Processing Systems}, 2024.

\bibitem[Snell et~al.(2024)Snell, Lee, Xu, and Kumar]{snell2024scaling}
Charlie Snell, Jaehoon Lee, Kelvin Xu, and Aviral Kumar.
\newblock Scaling llm test-time compute optimally can be more effective than scaling model parameters.
\newblock \emph{arXiv preprint arXiv:2408.03314}, 2024.

\bibitem[Team et~al.(2025)Team, Bai, Bao, Chen, Chen, Chen, Chen, Chen, Chen, Chen, et~al.]{team2025kimi}
Kimi Team, Yifan Bai, Yiping Bao, Guanduo Chen, Jiahao Chen, Ningxin Chen, Ruijue Chen, Yanru Chen, Yuankun Chen, Yutian Chen, et~al.
\newblock Kimi k2: Open agentic intelligence.
\newblock \emph{arXiv preprint arXiv:2507.20534}, 2025.

\bibitem[Wei et~al.(2025)Wei, Duchenne, Copet, Carbonneaux, Zhang, Fried, Synnaeve, Singh, and Wang]{wei2025swe}
Yuxiang Wei, Olivier Duchenne, Jade Copet, Quentin Carbonneaux, Lingming Zhang, Daniel Fried, Gabriel Synnaeve, Rishabh Singh, and Sida~I Wang.
\newblock Swe-rl: Advancing llm reasoning via reinforcement learning on open software evolution.
\newblock \emph{arXiv preprint arXiv:2502.18449}, 2025.

\bibitem[Wu et~al.(2025)Wu, Zhang, Xue, Liu, Diao, Zhu, Luo, Han, and Xie]{wu2025fastdllmtrainingfreeaccelerationdiffusion}
Chengyue Wu, Hao Zhang, Shuchen Xue, Zhijian Liu, Shizhe Diao, Ligeng Zhu, Ping Luo, Song Han, and Enze Xie.
\newblock Fast-dllm: Training-free acceleration of diffusion llm by enabling kv cache and parallel decoding, 2025.
\newblock URL \url{https://arxiv.org/abs/2505.22618}.

\bibitem[Xie et~al.(2025)Xie, Chen, Mao, Xu, Kong, et~al.]{xie2025teaching}
Zhihui Xie, Liyu Chen, Weichao Mao, Jingjing Xu, Lingpeng Kong, et~al.
\newblock Teaching language models to critique via reinforcement learning.
\newblock \emph{arXiv preprint arXiv:2502.03492}, 2025.

\bibitem[Xu et~al.(2025)Xu, Liu, Yin, Zhou, and Poovendran]{xu2025kodcode}
Zhangchen Xu, Yang Liu, Yueqin Yin, Mingyuan Zhou, and Radha Poovendran.
\newblock Kodcode: A diverse, challenging, and verifiable synthetic dataset for coding.
\newblock \emph{arXiv preprint arXiv:2503.02951}, 2025.

\bibitem[Ye et~al.(2025{\natexlab{a}})Ye, Gao, Gong, Zheng, Jiang, Li, and Kong]{ye2024beyond}
Jiacheng Ye, Jiahui Gao, Shansan Gong, Lin Zheng, Xin Jiang, Zhenguo Li, and Lingpeng Kong.
\newblock Beyond autoregression: Discrete diffusion for complex reasoning and planning.
\newblock \emph{International Conference on Learning Representations}, 2025{\natexlab{a}}.

\bibitem[Ye et~al.(2025{\natexlab{b}})Ye, Xie, Zheng, Gao, Wu, Jiang, Li, and Kong]{dream2025}
Jiacheng Ye, Zhihui Xie, Lin Zheng, Jiahui Gao, Zirui Wu, Xin Jiang, Zhenguo Li, and Lingpeng Kong.
\newblock Dream 7b, 2025{\natexlab{b}}.
\newblock URL \url{https://hkunlp.github.io/blog/2025/dream}.

\bibitem[Yu et~al.(2025)Yu, Zhang, Zhu, Yuan, Zuo, Yue, Dai, Fan, Liu, Liu, et~al.]{yu2025dapo}
Qiying Yu, Zheng Zhang, Ruofei Zhu, Yufeng Yuan, Xiaochen Zuo, Yu~Yue, Weinan Dai, Tiantian Fan, Gaohong Liu, Lingjun Liu, et~al.
\newblock Dapo: An open-source llm reinforcement learning system at scale.
\newblock \emph{arXiv preprint arXiv:2503.14476}, 2025.

\bibitem[Zellers et~al.(2019)Zellers, Holtzman, Bisk, Farhadi, and Choi]{zellers2019hellaswag}
Rowan Zellers, Ari Holtzman, Yonatan Bisk, Ali Farhadi, and Yejin Choi.
\newblock Hellaswag: Can a machine really finish your sentence?
\newblock \emph{arXiv preprint arXiv:1905.07830}, 2019.

\bibitem[Zheng et~al.(2023)Zheng, Yuan, Yu, and Kong]{zheng2023reparameterized}
Lin Zheng, Jianbo Yuan, Lei Yu, and Lingpeng Kong.
\newblock A reparameterized discrete diffusion model for text generation.
\newblock \emph{arXiv preprint arXiv:2302.05737}, 2023.

\bibitem[Zhuo et~al.(2024)Zhuo, Vu, Chim, Hu, Yu, Widyasari, Yusuf, Zhan, He, Paul, et~al.]{zhuo2024bigcodebench}
Terry~Yue Zhuo, Minh~Chien Vu, Jenny Chim, Han Hu, Wenhao Yu, Ratnadira Widyasari, Imam Nur~Bani Yusuf, Haolan Zhan, Junda He, Indraneil Paul, et~al.
\newblock Bigcodebench: Benchmarking code generation with diverse function calls and complex instructions.
\newblock \emph{arXiv preprint arXiv:2406.15877}, 2024.

\end{thebibliography}

\end{document}